# On Accurate and Reliable Anomaly Detection for Gas Turbine Combustors: A Deep Learning Approach

Weizhong Yan[1] and Lijie Yu[2]

[1]*General Electric Global Research Center, Niskayuna, New York 12309, USA*
yan@ge.com

[2]*General Electric Power & Water Engineering, Atlanta, Georgia 30339, USA*
Lijie.Yu@ge.com

## ABSTRACT

Monitoring gas turbine combustors' health, in particular, early detecting abnormal behaviors and incipient faults, is critical in ensuring gas turbines operating efficiently and in preventing costly unplanned maintenance. One popular means of detecting combustors' abnormalities is through continuously monitoring exhaust gas temperature profiles. Over the years many anomaly detection technologies have been explored for detecting combustor faults, however, the performance (detection rate) of anomaly detection solutions fielded is still inadequate. Advanced technologies that can improve detection performance are in great need. Aiming for improving anomaly detection performance, in this paper we introduce recently-developed deep learning (DL) in machine learning into the combustors' anomaly detection application. Specifically, we use deep learning to hierarchically learn features from the sensor measurements of exhaust gas temperatures. And we then use the learned features as the input to a neural network classifier for performing combustor anomaly detection. Since such deep learned features potentially better capture complex relations among all sensor measurements and the underlying combustors' behavior than handcrafted features do, we expect the learned features can lead to a more accurate and robust anomaly detection. Using the data collected from a real-world gas turbine combustion system, we demonstrated that the proposed deep learning based anomaly detection significantly indeed improved combustors' anomaly detection performance.

Deep learning, one of the breakthrough technologies in machine learning, has attracted tremendous research interests in recent years in the domains such as computer vision, speech recognition and natural language processing.

Deep learning, to the best of our knowledge, has not been used for any PHM applications, however. It is our hope that our initial work presented in this paper would shed some light on how deep learning as an advanced machine learning technology can benefit PHM applications and, more importantly, can stimulate more research interests in our PHM community.

## 1. INTRODUCTION

A combustion system is a critical component of gas turbines that burns fuel air mixture to create thrust or power. A heavy-duty industrial combustor typically operates under high temperature and high flow rate conditions that introduce significant thermodynamic stress to combustor components. Imbalanced fuel distribution and combustion instabilities are the main causes of different combustors' abnormalities, including fuel nozzle faults, liner cracks, transition piece defects, excessive vibration due to acoustic waves and heat release oscillations, and non-compliant emissions [Allegorico & Mantini (2014)]. Those abnormalities, if not detected early, could lead to catastrophic combustor failures or lean blowout, which trigger turbine trips; those abnormalities could also adversely affect the life of hot gas path components, or result in higher NOx and CO emissions. Consequently, reliably detecting abnormal behaviors and incipient faults earlier is important in ensuring gas turbines operating efficiently and in preventing costly turbine trips.

Combustor anomaly detection is technically challenging because gas turbine combustors are an extremely complex system, of which the operating conditions are heavily dependent on many factors, such as, machine type, fuel used, ambient conditions, and equipment aging.

Monitoring the exhaust gas temperatures measured at the gas turbine exhaust section is a popular means for detecting the combustor abnormalities [Allegorico & Mantini (2014)]. Exhaust temperature profiles provide valuable information about thermal performance of gas turbines and combustors, thus can be indicative to combustor health conditions.

Traditionally, for combustor anomaly detection, knowledge-based rules are applied to the exhaust temperature profiles. Such knowledge-based rules not only have inadequate detection performance (detection rate and false alarm rate), but also are laborious in designing and developing the rules. Aiming for more accurate and robust detection of combustors' incipient faults, thus for reducing unplanned downtimes and operation costs, in recent years we at GE have been pursuing advancing our anomaly detection technologies from the traditional knowledge-based rules to knowledge-augmented data-driven approaches. Specifically for combustor anomaly detection, we have explored different data-driven, machine learning technologies, such as SVM, random forests, and neural networks. Using advanced machine learning modeling techniques has made certain degree of improvement in detection performance, but not as significantly as we would like. We observed that it is the *feature engineering*, a process of extracting appropriate features or signatures from raw sensor measurements, which made bigger difference in combustors' detection performance.

In our early work we handcrafted a set of features based on domain and engineering knowledge of gas turbine combustors. Using such handcrafted features for our anomaly detection models yielded better detection performance than directly using raw exhaust temperatures for combustors' anomaly detection problems; however, handcrafting features is a manual process that is very much problem-specific and un-scalable. Thus it would be of great value if somehow we can automate the feature generation process. Deep learning (DL) is a sub-field of machine learning that involves learning good representations of data through multiple levels of abstraction. By hierarchically learning features layer by layer, with higher-level features representing more abstract aspects of the data, deep learning can discover sophisticated underlying structure and features. In recent years deep learning has attracted tremendous research attention and proven outstanding performance in many applications including image and video classification, computer vision, speech recognition, natural language processing, and audio recognition [Arel et al. (2010)].

Inspired by the success of deep learning in many other domains, in this paper we explore how deep learning can benefit PHM applications in general and combustor anomaly detection applications in particular. Broadly speaking deep learning has two types: supervised and unsupervised. Unsupervised feature learning, i.e., using unlabeled data to learn features, is the key idea behind the self-taught learning framework [Raina et al. (2007)]. Unsupervised feature learning is well suited for machinery anomaly detection since for PHM applications abundant unlabeled data are available and easily accessible, while accurately labeling industrial data is costly and, often time, impossible due to uncertainty of true events.

Deep learning, to the best of our knowledge, has not been used for any PHM applications, despite its success in many other domains. Our initial work presented in this paper can hopefully shed some light on how deep learning, as an advanced machine learning technology, can benefit PHM applications and, more importantly, our work here can hopefully stimulate more research interests in our PHM community.

The remaining of the paper is organized as follows. Section 2 provides related work on both anomaly detection and feature engineering & feature learning as well. We then give details on our methodology of using deep learning for combustor anomaly detection in Section 3. Use case study and its results are given in Section 4. We conclude our paper in Section 5.

## 2. RELATED WORK

### 2.1. Anomaly detection

Anomaly detection, a technique for finding patterns in data that do not conform to expected behavior, has been extensively used in a wide range of applications, such as fraud detection in credit card and insurance industries, intrusion detection in cyber-security industry, fault detection in industrial analytics, to name a few. Survey papers, for example, Chandola et al. (2009), provide a comprehensive review of different anomaly detection methods and applications.

Anomaly detection has been actively applied to different PHM applications including: aircraft engine fault detection [Tolani et al. (2006)], wind turbine fault detection [Zaher et al. (2009)], locomotive engine fault detection [Xue & Yan (2007)], marine gas turbine engine [Ogbonnaya et al. (2012)], and combined cycle power plants [Arranz et al. (2008)], to name a few.

There are a few studies specifically on combustor anomaly detection. For example, Mukhopadhyay and Ray (2013) used symbolic time series analysis for detecting lean blow-out in gas turbine combustors. The time series data they used for analysis were optical sensor data from the photomultiplier tube (PMT). In the work by Chakraborty et al (2008), the tailpipe wall friction coefficient was proposed as the failure precursor to flame out of thermal pulse combustors and several data-driven techniques (information theory, symbolic dynamics and statistical pattern recognition) were applied to pressure oscillation signals for estimating the friction coefficient of the tailpipe wall. One work that mostly relates to our study in this paper is by Allegorico and Mantini (2014). Similar to ours, they also performed combustor anomaly detection based on exhaust temperature thermocouples. They formulated the anomaly detection as a classification problem and used traditional neural networks and logistic regression as the classifiers. However, they didn't do any feature engineering to extract features. Rather they directly used the exhaust temperature profile as the inputs to classifier, which showed a reasonable detection performance on the small dataset the authors picked, but may not generalize well in real applications.

## 2.2. Feature engineering

Feature engineering is the process of transforming raw data into features that better represent the underlying problem to the predictive models, resulting in improved model accuracy on unseen data [Brownlee (2014)]. Feature engineering is arguably a critically important task in developing predictive solutions [Domingos (2012)]; and at the same time it is also a challenging but the least well-studied topic in machine learning and data-mining [Brownlee (2014)]. That is because feature engineering is a very much problem-specific, manual process that is typically performed by machine learning experts in conjunction with domain experts.

As features are highly application dependent, there is almost no universal feature set that works well for all applications. Over the years, though, many application domains do have developed a number of application-specific features that are popularly used. For example, frequency of each word in the bag-of-words for document classification, scale-invariant feature transform (SIFT) for object recognition [Lowe (1999)], and Mel-frequency cepstral coefficients (MFCC) for speech recognition [Davis and Mermelstein (1980)], and defect frequencies for vibration analysis. These commonly used features serve as a good starting point for feature engineering.

In literature, publications specific on feature engineering are very sparse as stated by Brownlee (2014) that "feature engineering is another topic which doesn't seem to merit any review papers or books, or even chapters in books". Recently there are a few attempts on developing feature engineering tools that aim for facilitating the feature engineering task. For example, Anderson et al. (2014) proposed a feature engineering development environment that allows the user to write feature engineering code and evaluate the effectiveness of the engineered features. Heimerl et al. (2012) developed FeatureForge tool that uses interactive visualization for supporting feature engineering for natural language processing.

Feature engineering for PHM applications also attracts researchers' attention. For example, Yan et al. (2008) provided a survey on feature extraction for bearing PHM applications.

## 2.3. Feature (representation) learning

Feature learning, also called representation learning, is a sub-field of machine learning where the focus is to learn a transformation of raw data input to a representation that can be effectively exploited in machine learning tasks. Feature learning becomes an active research topic in recent years as deep learning or deep representation learning becomes a hot research topic [NIPS (2014), ICML (2013), and ICLR (2015)]. Deep representation learning has created great impact in the areas such as speech recognition [Deng, et al. (2010)], object recognition [Hinton, et al. (2006)], and natural language processing [Collobert, et al. (2011)]. Deep representation learning employs deep learning architecture for feature learning. By stacking up multiple layers of shallow leaning blocks, higher layer features learned from lower layer features represent more abstract aspects of the data, and thus can be more robust to variations.

Feature learning can be broadly categorized into unsupervised and supervised learning groups [Wikipedia (2015)]. Supervised representation learning includes primarily the traditional multi-layer neural networks and supervised dictionary learning. Unsupervised representation learning, a key idea behind the self-taught learning framework [Raina et al. (2007)], covers more techniques, ranging from traditional methods such as PCA, ICA, and k-means, to advanced methods such as autoencoders, RBM, and sparse coding. Unsupervised representation learning has several advantages. For example, the explicitly learned features can be used for different prediction models. Unsupervised representation learning can also be an important component of transfer learning [Bengio (2011)]. Successful feature learning algorithms and their applications can be found in recent literature using a variety of approaches, such as RBMs [Hinton et al. (2006)], autoencoders [Hinton & Salakhutdinov (2006)], sparse coding [Lee et al. (2007)], and K-means [Coates et al. (2011))]. The most popular building blocks include autoencoder and restricted Boltzmann machines (RBM). Denosing autoencoders (DAE), a variant of classic autoencoders, and its deep counterpart, stacked denoising autoencoders (SDAE) [Vincent, et al. (2010)], have been used as a representation learning algorithm for several applications, for example, for pose-based action recognition [Budiman, et al. (2014)], for tag recommendation [Wang, et al. (2015)], and for handwritten digits recognition [Vincent, et al. (2010)]. SDAE has not been used for PHM applications, however.

## 3. METHODOLOGY

For combustor anomaly detection concerned in this paper, we adopt unsupervised representation learning scheme. Under this scheme, features are explicitly learned un-supervisingly (without class labels) and the explicitly learned features are then used as input for a separate supervised model (classifier). There are different shallow learning blocks that can be stacked up to form deep feature learning structures. For combustor anomaly detection concerned in this paper we adopt the SDAE proposed by Vincent, et al. in 2010 as the unsupervised representation learning algorithm, which has the denoising autoencoder (DAE), a variant of autoencoder (AE), as its shallow learning blocks. The main reason we chosen SDAE is that denoising autoencoders can learn features that are more robust to input noise and thus useful for classification. The features learned from the SDAE are then taken as the input to a separate NN classifier, extreme learning machine (ELM), for anomaly detection. Figure 1 illustrates both the SDAE for deep feature learning and the ELM for classification for combustor anomaly detection. Both SDAE and ELM are described in detail as follows.

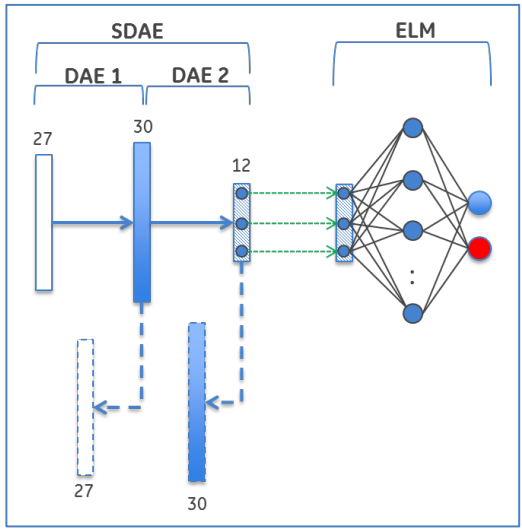

Figure 1: Overall structure of unsupervised feature learning for combustor anomaly detection

### 3.1. SDAE for unsupervised feature learning

Stacked denoising autoencoder (SDAE), introduced by Vincent et al (2010), is a deep learning structure that has denoising autoencoder (DAE) as its shallow learning blocks. DAE is a variant of classic autoencoder (AE). While details can be found in many references, we provide a brief description of AE and DAE as follows.

An auto-encoder (AE), in its basic form, has two parts: an encoder and a decoder. The encoder is a function that maps an input $x \in \Re^{d_x}$ to hidden representation $h(x) \in \Re^{d_h}$, that is, $h(x) = s_f(Wx + b_h)$, where $s_f$ is a nonlinear activation function, typically a logistic sigmoid function. The decoder function maps hidden representation $h$ back to a reconstruction y: $y = s_g(W'h + b_y)$, where $s_g$ is the decoder's activation function, typically either the identity function (yielding linear reconstruction) or a sigmoid function.

Autoencoder training consists of finding parameters $\theta = \{W, b_h, b_y\}$ that minimize the reconstruction error on a training set of examples, $D$. That is: $J_{AE}(\theta) = \sum_{x \in D} L(x, g(f(x)))$, where L is the reconstruction error.

The reconstruction error, L, can be the squared error $L(x, y) = -\sum_{i=1}^{d_x}(x_i - y_i)^2$ when $s_g$ is linear; or the cross-entropy loss $L(x, y) = -\sum_{i=1}^{d_x} x_i log(y_i) + (1 - x_i)log(1 - y_i)$ when $s_g$ is the sigmoid.

To prevent autoencoders from learn the identity function that has zero reconstruction errors for all inputs, but does not capture the structure of the data-generating distribution, it is important that certain regularization is needed in the training criterion or the parametrization. A particular form of regularization consists in constraining the code to have a low dimension, and this is what the classical auto-encoder or PCA do.

The simplest form of regularization is weigh-decay which favors small weights by optimizing the following cost function: $J_{AE+wd}(\theta) = \sum_{x \in D} L(x, g(f(x))) + \lambda \sum_{ij} W_{ij}^2$

Another form of regularization is by corrupting input x during training the autoencoder. Specifically, corrupting the input x in the encoding step, but still to reconstruct the clean version of x in the decoding step. This is called denoising autoencoder (DAE). The goal here is not for denoising of input signals per se. Rather denoising is advocated as a training criterion such that the extracted features will constitute better high-level representation.

Vincent et al (2010) discussed three ways to corrupt inputs: 1) additive isotropic Gaussian noise: $\hat{x}|x \sim \aleph(x, \sigma^2 I)$; 2) masking noise: a fraction n of the elements of x (chosen at random for each example) is forced to 0; and 3) salt-and-pepper noise: a fraction n of the elements of x (chosen at random for each example) is set to their minimum or maximum possible value (typically 0 or 1) according to a fair coin flip. While the additive Gaussian noise is a natural choice for real valued inputs, the salt-and-pepper noise is a natural choice for input domains which are interpretable as binary or near binary such as black and white images or the representations produced at the hidden layer after a sigmoid squashing function. The masking noise is equivalent to turning off components that have missing values. Thus DAE is trained to fill-in the missing data, which forces the extracted features to better capture the dependence among the all input variables.

### 3.2. ELM for classification

For combustor anomaly detection problem concerned in this paper, we use SDAE to learn features, which are then used as the input to the extreme learning machine (ELM) classifier. ELM is a special type of feedforward neural networks introduced by Huang, et al. [Huang et al. (2006)]. Unlike in other feedforward neural networks where training the network involves finding all connection weights and bias, in ELM, connections between input and hidden neurons are randomly generated and fixed, that is, they do not need to be trained; thus training an ELM becomes finding connections between hidden and output neurons only, which is simply a linear least squares problem whose solution can be directly generated by the generalized inverse of the hidden layer output matrix [Huang et al. (2006)]. Because of such special design of the network, ELM training becomes very fast. ELM has one design parameter, i.e., the number of hidden neurons. Studies have shown that the ELM prediction performance is not too sensitive to the design parameter, as long as it is large enough, say 1000, which simplifies ELM design and makes ELM a more attractive model. Numerous empirical studies and recently some analytical studies as well have shown that ELM is an efficient and effective model for both classification and regression [Huang, et al. (2012)]. Once again the ELM classifier takes feature learned from SDAE as the inputs and outputs probabilities of abnormality for gas turbine combustors.

## 4. CASE STUDY AND RESULTS

### 4.1. The business application

The asset of interest for our study is a combustor assembly used in heavy-duty industrial gas turbines. The combustor assembly is composed of a plural of individual combustion chambers. Fuel and compressed airflow is mixed and combusted in each combustion chamber, then the expanded hot gas is guided through a hot gas path along a number of turbine stages to derive work. A number of thermal couples (TC) are arranged in the turbine exhaust system to measure the exhaust gas temperature. The number of TC measurements varies depending on the turbine frames being monitored. It is a standard practice using TC temperature profile to infer combustor health condition. A typical TC temperature profile after mean normalization is shown in Figure 2.

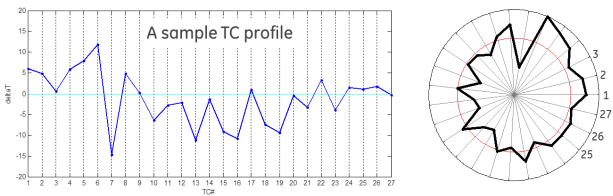

Figure 2: A sample TC profile

### 4.2. Data description

Our database has several years of data sampled at once-per-minute. For demonstration purpose, in this study, we use several months of data for one turbine. Specifically, we use three months of event-free data and four months of data where 10 events occurred somewhere in the four-month window. After filtering out bad data points and those data points corresponding to part load condition (TNH<95%), we end up with 13,791 samples before the POD events (these samples are event-free and are considered to be normal), 300 samples for the POD events, and 47,575 samples after the POD events. The number of thermocouples for this turbine is 27, which is equal to the number of combustor cans.

In this study, we treat the 13,791 samples before the POD events as event-free (normal) data for unsupervised feature learning. And we use the rest of data (both POD events and event-free data) for training and testing the classifier.

### 4.3. Model design

For unsupervised feature learning, we use 2-layer SDAE. While DAE1 has 30 hidden neurons, DAE2 has 12 hidden neurons (See Figure 1). Activation functions for hidden neurons of both DAEs are sigmoid function. The noise rate is 0.2. The learning rate and momentum are 0.02 and 0.5, respectively. The number of epochs for learning is 200 for both DAEs. DAEs are implemented in Matlab R2014a.

The ELM classifier, as discussed in Section 3, has one design parameter, that is, the number of hidden neurons. Generally setting the number of hidden neurons to a large number, i.e., 1000 in this study.

As described in the previous section, our data is highly imbalanced between normal and abnormal classes (with majority-to-minority ratio of approximately 150), which deserves a special attention in classifier modeling. In literature there are many different strategies handling imbalanced data. He and Garcia (2009) provided a comprehensive review of different imbalance learning strategies. In this paper we take advantage of ELM's capability of weighting samples during learning.

### 4.4. Results

To demonstrate effectiveness of unsupervised feature learning for combustor anomaly detection, we compare classification performance between using the learned features and using knowledge-driven, handcrafted features. Remember we use the identical setting of the ELM classifier for the comparison. In other word, using different feature sets is the only difference between the two designs in comparing classification performance. We use ROC curves as the classification performance measure for comparison. We employ 5-fold cross-validation for model training and validation. To obtain more robust comparison we run the 5-fold cross-validation 10 times, each time with different randomly splitting of 5 folds of the data.

The handcrafted features are primarily simple statistics calculated on TC profiles. These simple statistics essentially capture engineering knowledge of combustor TC profiles associated with different combustor states (healthy and fault). The 12 handcrafted features are listed in Table 1 below.

**Table 1 - Handcrafted Features**

| ID | Feature | Description |
|----|---------|-------------|
| 1  | DWATT   | Raw turbine load |
| 2  | TNH     | Raw turbine speed |
| 3  | MAX     | Max TCs |
| 4  | MEN     | Mean TCs |
| 5  | STD     | Standard deviation of TCs |
| 6  | MED     | Median of TCs |
| 7  | DIF     | # diff b/w positive & negative TCs |
| 8  | ZR      | Zero crossing |
| 9  | KR      | kurtosis |
| 10 | SK      | skewness |
| 11 | M3S     | Max of 3-pt sum |
| 12 | M3M     | Max of 3-pt median |

The ROCs for the 10 runs of 5-fold cross-validation using the handcrafted features are shown *in blue* in Figure 3.

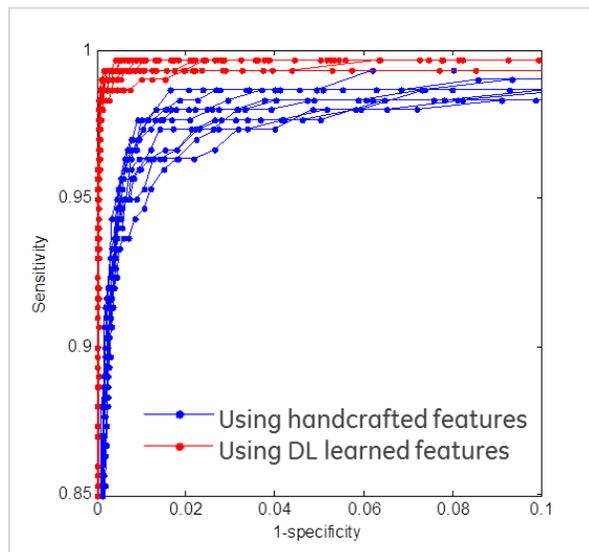

Figure 3: ROCs comparison

The 12 learned features are shown in Figure 4. Unlike the handcrafted features, each of which is a numerical number, the learned features are patterns representing the data (TC profiles) underlying structures. As the result, the learned features are more powerful in representing the data, thus performing better in classification. The ROCs for the 10 runs of 5-fold cross-validation using the learned features are shown *in red* in Figure 3.

From the ROC comparison in Figure 3, one can see clearly that the deep learned features give significant better classification performance than the handcrafted features do. Also from the ROCs one can see that using the deep learned features yields smaller variation in ROCs than using the handcrafted features. For example, when false positive rate (1-specifty) is at 1%, the mean and the standard deviation of the true positive rates (sensitivity) for both the deep learned features and the handcrafted features are approximately $0.99 \pm 0.01$ and $0.96 \pm 0.02$, respectively.

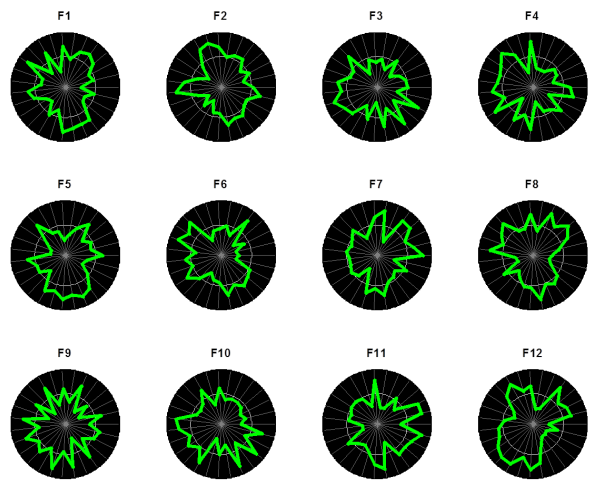

Figure 4: The 12 learned features

### 5. CONCLUSION

Accurately detecting gas turbine combustor abnormalities is important in reducing O&M costs of power plants. Traditional rule-based anomaly detection solutions are inadequate in achieving the desired detection performance. Adopting more advanced machine learning technologies as a means of improving combustors' detection performance is in great need. Realizing that generating good features is both a critically important and challenging task in developing machine learning solutions, in this paper we attempt to leverage recently developed unsupervised representation learning, a key part of deep learning, for finding more salient features from raw TC measurements for achieving more accurate and robust combustor anomaly detection. More specifically, we want to know if representation learning, which has approved to be effective in many other applications, can be an effective feature generation means for PHM applications. By applying SDAE we demonstrated that deep feature learning could effectively generate features from the raw time-series TC measurements, which thus improved combustor anomaly detection.

Unsupervised representation learning, or deep learning in general, has proven to be an effective ML technology in other domains, but has not been used for any PHM applications. It is our hope that our initial work presented in this paper can shed some light on how deep learning as an advanced machine learning technology can benefit PHM applications and stimulate more research interests in our PHM community. In future we would like to conduct more thorough studies of combustor anomaly detection by using more real-world data. We would also like to explore other deep learning methods other than SDAE for combustor anomaly detection and other PHM applications as well.

**ACKNOWLEDGEMENT**

We would like to thank one of our colleagues, Dr. Johan Reimann, for insightful discussion of deep learning technology over the course of this study.

**BIOGRAPHIES**

**Weizhong Yan, PhD, PE,** has been with the General Electric Company since 1998. Currently he is a Principal Scientist in the Machine Learning Lab of GE Global Research Center, Niskayuna, NY. He obtained his PhD from Rensselaer Polytechnic Institute, Troy, New York. His research interests include neural networks (shallow and deep), big data analytics, feature engineering & feature learning, ensemble learning, and time series forecasting. His specialties include applying advanced data-driven analytic techniques to anomaly detection, diagnostics, and prognostics & health management of industrial assets such as jet engines, gas turbines, and oil & gas equipment. He has authored over 70 publications in referred journals and conference proceedings and has filed over 30 US patents. He is an Editor of International Journal of Artificial Intelligence and an Editorial Board member of International Journal of Prognostics and Health Management. He is a Senior Member of IEEE.
.

**Lijie Yu, PhD,** has joined GE since 2011. She started her career at the GE Global Research Center at Niskayuna, NY as a Computer Scientist. During that time, she built experience developing advanced algorithms for intelligent systems across multiple GE businesses. She is specialized in signal processing, machine learning, information fusion, and equipment health management. Currently she is senior analytics engineer at GE Power and Water, supporting power generation equipment remote monitoring and fleet management. She is holder of 7 awarded patents and authored 15 publications. She is also a certified Black Belt.